\documentclass[10pt,twocolumn,letterpaper]{article}

\usepackage[pagenumbers]{cvpr} 

\usepackage[pagebackref,breaklinks,colorlinks,allcolors=cvprblue]{hyperref}
\usepackage{multirow}
\usepackage[table]{xcolor} 
\usepackage{amsmath}
\usepackage{amssymb}
\usepackage{lineno}
\usepackage{float}
\usepackage{stfloats}
\usepackage{bm}
\usepackage{algorithm}
\usepackage{algpseudocode}

\newcommand{\method}{{\sc CamVerse}}
\newcommand{\myPara}[1]{\vspace{.05in}\noindent\textbf{#1}}
\definecolor{cvprblue}{rgb}{0.21,0.49,0.74}
\definecolor{blue}{RGB}{82, 115, 197}
\definecolor{green}{RGB}{133, 187, 84}
\definecolor{red}{RGB}{176, 36, 24}
\definecolor{ourblue}{rgb}{0.5, 0.6, 0.7}

\def\paperID{1839} 
\def\confName{CVPR}
\def\confYear{2026}

\title{Taming Camera-Controlled Video Generation with Verifiable Geometry Reward}

\author{
Zhaoqing Wang\textsuperscript{1} \quad
Xiaobo Xia\textsuperscript{2}\thanks{Corresponding author.} \quad
Zhuolin Bie\textsuperscript{1} \quad
Jinlin Liu\textsuperscript{1} \\[0.5em]
Dongdong Yu\textsuperscript{1} \quad
Jia-Wang Bian\textsuperscript{3} \quad
Changhu Wang\textsuperscript{1}$^*$\\[0.5em]
\small
\textsuperscript{1}AIsphere \quad
\textsuperscript{2}National University of Singapore \quad
\textsuperscript{3}Nanyang Technological University \\[0.5em]
{\tt\small xiaoboxia.uni@gmail.com \quad wangchanghu@aishi.ai}
}

\begin{document}

\twocolumn[{
\renewcommand\twocolumn[1][]{#1}
\maketitle
\begin{center}
    \centering
    \captionsetup{type=figure}
    \includegraphics[width=1.0\textwidth]{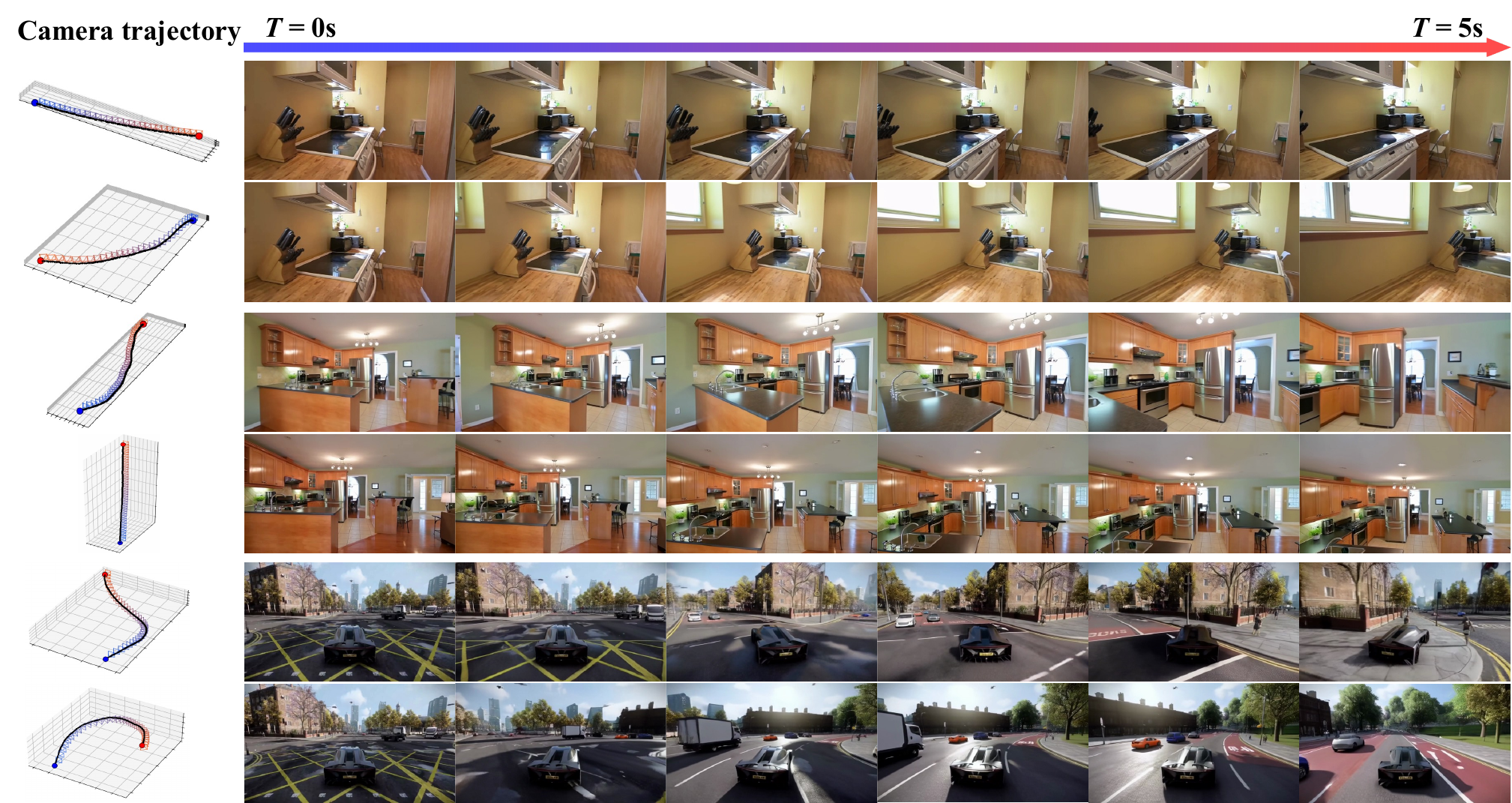}
    \caption{
        \textbf{Camera-controlled video generation with \method.}
        Given an initial frame, a text prompt, and a user-specified 3D camera trajectory, our model synthesizes a video that follows the trajectory while preserving scene layout and appearance.
        \textit{Left}: input trajectories (start in blue, end in red).
        \textit{Right}: frames sampled from $T=0$ to $T=5$s.
        Rows are grouped in pairs that share the same first frame and prompt; only the camera path differs within each pair. These examples span diverse camera movements (\eg, planar curves and out-of-plane arcs) and scenarios (\eg, indoor kitchens and outdoor driving), showcasing precise camera control.
        }
    \label{fig:teaser}
\end{center}
}]

\begin{abstract}
Recent advances in video diffusion models have remarkably improved camera-controlled video generation, but most methods rely solely on supervised fine-tuning (SFT), leaving online reinforcement learning (RL) post-training largely underexplored.
In this work, we introduce an online RL post-training framework that optimizes a pretrained video generator for precise camera control.
To make RL effective in this setting, we design a verifiable geometry reward that delivers dense segment-level feedback to guide model optimization. Specifically, we estimate the 3D camera trajectories for both generated and reference videos, divide each trajectory into short segments, and compute segment-wise relative poses. The reward function then compares each generated-reference segment pair and assigns an alignment score as the reward signal, which helps alleviate reward sparsity and improve optimization efficiency. Moreover, we construct a comprehensive dataset featuring diverse large-amplitude camera motions and scenes with varied subject dynamics. Extensive experiments show that our online RL post-training clearly outperforms SFT baselines across multiple aspects, including camera-control accuracy, geometric consistency, and visual quality, demonstrating its superiority in advancing camera-controlled video generation.
\end{abstract}    
\section{Introduction}
\label{sec:intro}
Video diffusion models have advanced rapidly, delivering high-fidelity and temporally coherent frames directly from textual prompts~\cite{svd,sora,kling,wan2025wan,pixverse}. Beyond raw quality, many methods increasingly accept user-specified controls and conditioning signals, enabling flexible manipulation over content and motion~\cite{sparsectrl,xiaoyu_motioni2v,cameractrl,fu20243dtrajmaster,wang2025mmgen}. In parallel, action-conditioned models in interactive media show that policies can predict state transitions and future observations, pointing toward exploration of generated worlds rather than mere depiction~\cite{yu2025gamefactory,thematrix,parkerholder2024genie2,gamengen}. Within general video generation, camera control has emerged as a natural interface for such exploration by injecting camera parameters into pretrained diffusion models~\cite{camco,ac3d,motionctrl,bahmani2024vd3d,kuang2025collaborative,cavia}. These advances make it essential to reliably translate input camera trajectories into geometrically consistent videos, especially for film creation, environment simulation, and immersive experiences.

Despite their impressive progress, the predominant methods for camera-controlled video generation remain supervised fine-tuning~(SFT), while online reinforcement learning~(RL) is largely underexplored~\cite{liu2025flow,shao2024deepseekmath,xue2025dancegrpo,liu2025improving,liu2025videodpo}. This gap mainly persists for two reasons. First, it is nontrivial to design verifiable and geometry-aware reward functions for high-dimensional video outputs, and even estimating reliable camera trajectories from generated frames can be challenging. Second, camera trajectory is intrinsically continuous, yet most rewards used in AI-generated content~(AIGC) are instance-level (\eg, image/clip scores), resulting in sparse feedback and inefficient credit assignment during optimization. Nevertheless, online RL retains substantial untapped potential in this setting. Unlike SFT, which passively fits to fixed training data, online RL actively generates samples from the learned conditional distribution, reinforcing target camera motions through positive feedback while discouraging failure modes through negative feedback. Moreover, RL offers a principled mechanism to incorporate strong priors from large 3D models~\cite{wang2025vggt,wang2025pi,keetha2025mapanything} directly into video generators, thereby enhancing geometric consistency and improving camera-control accuracy of generated videos.

\begin{figure}[!tp]
    \centering
    \includegraphics[width=0.85\linewidth]{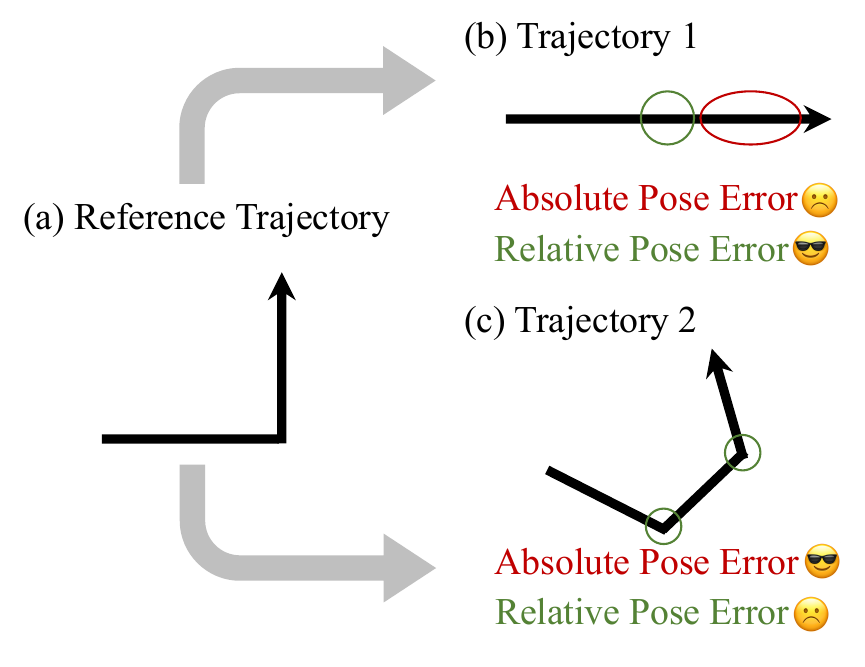}
    \vspace{-12pt}
    \caption{
        \textbf{Comparison between the absolute and relative pose error.}
        Given a reference trajectory, the model randomly generates two videos with corresponding trajectories. The absolute pose error prefers global-similar but locally incorrect results, while the relative pose error emphasizes locally consistent and smooth results. The latter is more aligned with the task characteristics.
    }
    \vspace{-9pt}
    \label{fig:intro}
\end{figure}

In this work, we introduce \method, an online RL post-training framework that optimizes a video generative model for precise camera control. At its core lies a verifiable geometry reward that translates camera-control accuracy into dense segment-level feedback. Specifically, we first estimate 3D camera trajectories for both the generated and the reference videos using a large 3D model. Each trajectory is then divided into short non-overlapping segments, within which we compute relative camera motion instead of absolute poses. Each generated reference segment pair is subsequently scored with an alignment metric, forming a dense reward signal. As illustrated in Figure~\ref{fig:intro}, the absolute pose error captures the global shape of trajectories and favors globally similar but locally inaccurate trajectories, whereas the relative pose error emphasizes local smoothness and continuity, which aligns with the behavior desired for camera-controlled video generation. Moreover, obtaining absolutely accurate videos typically requires lots of rollout, which is sample-inefficient for online RL.

Apart from that, most available datasets mainly focus on static video clips with camera parameters, which causes a negative effect on the motion strength of generated videos, and are challenging to encourage models to disentangle camera movement from subject dynamics. To overcome these limitations, we construct a large-scale dataset of video clips collected from both real-world footage and video games, emphasizing diverse and large-amplitude camera movement. We estimate camera trajectories with state-of-the-art 3D models~\cite{huang2025vipe,wang2025pi} and apply a multi-stage filtering pipeline to remove failure reconstruction cases and broaden the trajectory distribution. The resulting dataset contains 315k videos with annotated camera trajectories, of which 314k are used for training and 1k for testing. Before delving into details, our contributions are summarized as follows:
\begin{itemize}
    \item We propose \method, the first online RL post-training framework for camera-controlled video generation. The framework is equipped with a verifiable geometry reward that delivers dense segment-level feedback, effectively alleviating reward sparsity.
    \item We curate a new large-scale dataset spanning real-world footage and game environments, which is annotated with camera trajectory and covers diverse camera movements and varied subject dynamics.
    \item Extensive experiments demonstrate consistent improvements over SFT baselines in camera-control accuracy, geometric consistency, and visual quality~(\textit{cf.}, Figure~\ref{fig:teaser}).
\end{itemize}

\section{Related Work}
\label{sec:related_work}

\subsection{Video Diffusion Models}
Recent years have seen rapid progress in video generation, driven by advances in model architectures~\cite{ho2022vdm}, the availability of large-scale datasets~\cite{webvid10m,panda70m,HDVILA100M}, the introduction of comprehensive evaluation benchmarks~\cite{vbench,vbench++}, and improvements in training techniques~\cite{edm,recitfiedflow}.
Two major research directions in video generation are text-to-video (T2V) and image+text-to-video (IT2V).
The former directly synthesizes a video from a natural-language prompt.
Early approaches~\cite{alignyourlatents,svd,guo2023animatediff,gupta2024photorealistic,cogvideo} adapt UNet-based text-to-image (T2I) models by introducing temporal modules, which enable frame-to-frame dynamics while reusing strong image priors.
Subsequent works further scale data and computation to enhance visual fidelity and diversity~\cite{alignyourlatents,cogvideo,makeavideo,svd}.
Beyond pure text conditioning, IT2V methods~\cite{hu2022make,guo2023animatediff,xing2024dynamicrafter} take a still image with a caption as input to guide video synthesis.
More recent systems employ diffusion transformer architectures~\cite{dit,sd3,labs2025flux}, large-scale training pipelines, advanced post-training techniques~\cite{liu2025improving,liu2025videodpo} to achieve long-range temporal coherence and high visual quality, represented by models such as \cite{ma2024latte,sora,lin2024open,yang2024cogvideox,kong2024hunyuanvideo,kling,pixverse,yang2024cogvideox,noauthor_cosmos_nodate,wan2025wan,ma2025step}.
Despite this progress, most existing models still operate purely in 2D pixel space without explicit 3D supervision, often leading to geometric distortions under complex viewpoint changes.
Motivated by this limitation, we propose a post-training framework that explicitly rewards adherence to desired camera trajectories and multi-view geometric consistency, thereby improving both controllability and consistency in generated videos.

\subsection{Camera-Controled Video Generation}
Controlling camera movement in video generation is commonly achieved by injecting camera information into pretrained models. For example, MotionCtrl~\cite{motionctrl}, CameraCtrl~\cite{cameractrl}, and I2VControl\mbox{-}Camera~\cite{i2vcontrolcamera} condition on camera parameters (\eg, extrinsics, point trajectories, and Plücker embeddings~\cite{lfn}). Afterwards, CamCo~\cite{camco} incorporates epipolar constraints into attention, while CamTrol~\cite{camtrol} leverages 3D point-cloud information. AC3D~\cite{ac3d,bahmani2024vd3d} presents a deep analysis of camera-controlled video generation and designs an effective interface for injecting camera representations into pretrained backbones. Beyond single-camera settings, recent works~\cite{kuang2025collaborative,cavia,vividzoo,bai2024syncammaster} address multi-camera synchronization and cross-view consistency. CameraCtrl2~\cite{he2025cameractrl} further introduces a scalable data pipeline with camera-trajectory annotations for dynamic scenes, substantially improving the realism of dynamic content.
Different from previous works, we propose a reinforcement learning post-training framework with verifiable geometry reward. This directly optimizes the model to follow user-specified camera trajectories while maintaining 3D-consistent content across frames.

\subsection{Feed-Forward 3D Reconstruction}
Feed-forward models that regress 3D scene structure from image sets offer an efficient alternative to optimization-heavy pipelines. In more detail, Dust3R~\cite{wang2024dust3r} predicts pairwise point clouds in the first-view coordinate frame, but scaling to larger scenes typically requires a brittle global alignment. Fast3R~\cite{yang2025fast3r} improves scalability by performing joint inference over thousands of images, mitigating the need for costly post-hoc alignment. Subsequent work simplifies the problem via decomposition. FLARE~\cite{zhang2025flare} first estimates camera poses and then infers geometry, while VGGT~\cite{wang2025vggt} jointly predicts camera parameters and dense geometry using multi-task learning and large-scale data. These feed-forward methods still anchor predictions to a reference frame. To address that, $\pi^3$~\cite{wang2025pi} addresses this with a fully permutation-equivariant architecture that removes reference-frame bias. Leveraging these advances, we propose a novel post-training framework that injects powerful 3D prior learned by feed-forward models into camera-controlled video generation. This sets new state-of-the-art results across diverse scenes.
\section{Methodology}
\label{sec:method}

\subsection{Preliminary}
\label{sec:pre}
\myPara{Camera-controlled video generation.}
Let $\mathbf{x}_{1:N}=\{\mathbf{x}_n\}_{n=1}^{N}$ denote a video and $\mathbf{c}$ denote a text prompt. Each frame $n$ is accompanied by camera intrinsics $\mathbf{K}_n\in\mathbb{R}^{3\times3}$ and camera-to-world extrinsics $\mathbf{E}_n=[\mathbf{R}_n\mid \mathbf{t}_n]\in\mathrm{SE}(3)$ with $\mathbf{R}_n\in\mathrm{SO}(3)$ and $\mathbf{t}_n\in\mathbb{R}^3$. Directly feeding raw $(\mathbf{K}_n,\mathbf{R}_n,\mathbf{t}_n)$ to the model is poorly coupled with pixel-space features and leaves translations unconstrained. In contrast, we construct the per-frame Plücker embedding~\cite{lfn} as camera condition. Concretely, we compute the world-space ray direction $\mathbf{d}_{u,v}\propto\mathbf{R}_n\,\mathbf{K}_n^{-1}[u,v,1]^{\!\top}$, and the camera center $\mathbf{o}_n$ for each pixel $(u,v)$. The Plücker line coordinates for pixel $(u,v)$ are $\mathbf{p}_{u,v}=\big(\,\mathbf{o}_n\times \mathbf{d}_{u,v},\; \mathbf{d}_{u,v}\,\big)\in\mathbb{R}^{6}$ and normalized to the unit direction. Stacking all pixels yields a per-frame embedding $\mathbf{P}_n\in\mathbb{R}^{6\times h\times w}$ and the condition of a camera trajectory is $\mathbf{P}_{1:N}=\{\mathbf{P}_n\}_{n=1}^{N}$. Given a noisy latent $\mathbf{z}_t$, a diffusion timestep $t$ and these conditions, a pretrained video diffusion model $\pi_\theta$ is optimized with the standard flow-matching objective:
\begin{equation}
\mathcal{L}_{\mathrm{FM}}(\theta)
=\mathbb{E}_{\mathbf{z}_t,t,\mathbf{c},\mathbf{P}_{1:N}}
\!\left[\left\|\bm{\epsilon}-\pi_\theta(\mathbf{z}_t\mid t,\mathbf{c},\mathbf{P}_{1:N})\right\|_2^2\right],
\end{equation}
where $\bm{\epsilon}$ denotes the target velocity.

\begin{figure*}
    \centering
    \includegraphics[width=1\linewidth]{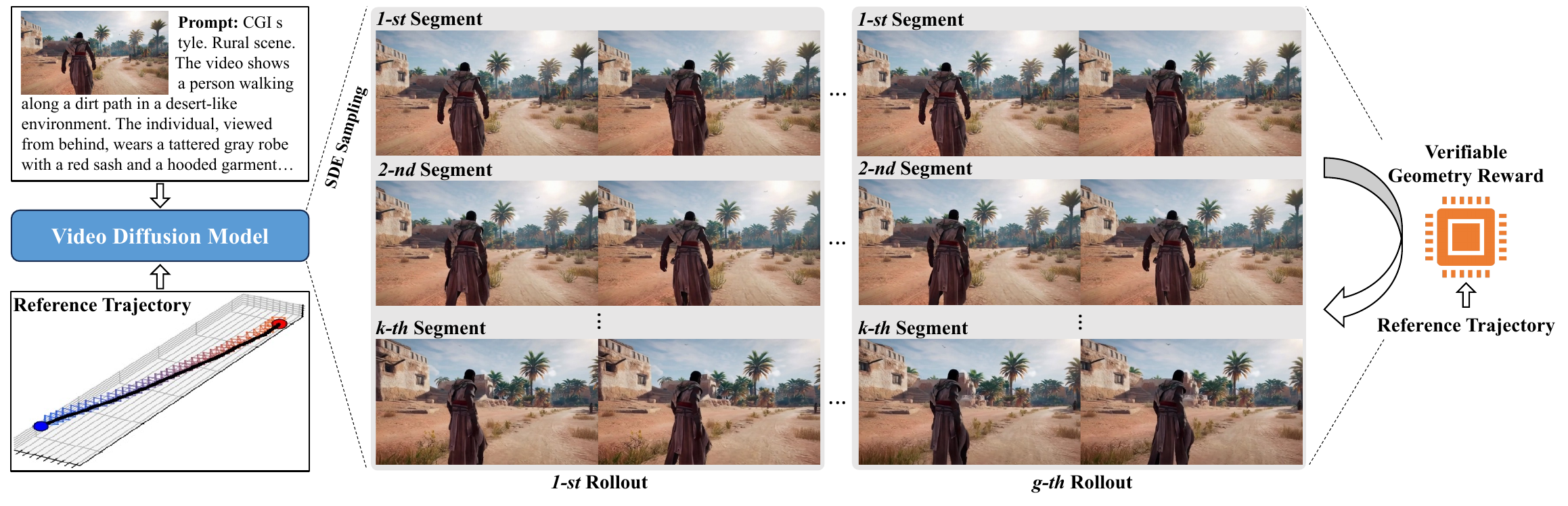}
    \caption{
    \textbf{Framework of \method.} Given the input first frame, a text prompt, and a reference camera trajectory, the video diffusion model $\pi_\theta$ samples $G$ rollouts by a stochastic reverse process (SDE sampling). Each rollout is partitioned into non-overlapping segments. For each rollout, a large 3D model is used to estimate the camera trajectory. After aligning the generated trajectory to the reference one, the verifiable geometry reward compares each pair of generated-reference segments in a relative manner and assigns a dense alignment score. With this type of reward signal, we conduct GRPO fine-tuning, improving camera-control accuracy and geometric consistency.    }
    \label{fig:method}
\end{figure*}

\myPara{Online RL.} 
We treat the video diffusion model as a stochastic policy that samples videos, $\mathbf{x}_{1:N} \sim \pi_{\theta}(\cdot \mid \mathbf{c}, \mathbf{P}_{1:N})$.
In each online iteration, given a pair of conditions $(\mathbf{c}, \mathbf{P}_{1:N})$, we draw $G$ rollouts from the current policy and evaluate each with a reward function $r(\mathbf{x}_{1:N}, \mathbf{c}, \mathbf{P}_{1:N})$. Afterwards, by omitting subscripts of $\mathbf{x}_{1:N}$ and $\mathbf{P}_{1:N}$ below, the policy $\pi_{\theta}$ is updated to increase the expected reward as:
\begin{equation}
\max_{\theta}\;
\mathbb{E}_{\mathbf{x}\sim \pi_{\theta}}
\!\left[ r(\mathbf{x}, \mathbf{c}, \mathbf{P}) \right],
\end{equation}
where $r(\cdot)$ denotes a reward function.
Interpreting the reward as an ``optimality'' probability 
\(r(\mathbf{x},\mathbf{c},\mathbf{P}_{1:N})\in[0,1]\),
a sample \(\mathbf{x}\) drawn from the old policy is assigned to a positive subset with probability \(r(\cdot)\) and to a negative subset with probability \(1-r(\cdot)\).
This induces two reweighted distributions:
\begin{equation}
\begin{aligned}
\pi^{+}(\mathbf{x}\mid \mathbf{c},\mathbf{P})
&=\frac{ r(\mathbf{x},\mathbf{c},\mathbf{P})\,
\pi_{\theta_{\text{old}}}(\mathbf{x}\mid \mathbf{c},\mathbf{P}) }
{ \mathbb{E}_{\mathbf{x}\sim \pi_{\theta_{\text{old}}}}
\!\left[r(\mathbf{x},\mathbf{c},\mathbf{P})\right] },\\
\pi^{-}(\mathbf{x}\mid \mathbf{c},\mathbf{P})
&=\frac{ \big(1-r(\mathbf{x},\mathbf{c},\mathbf{P})\big)\,
\pi_{\theta_{\text{old}}}(\mathbf{x}\mid \mathbf{c},\mathbf{P}) }
{ \mathbb{E}_{\mathbf{x}\sim \pi_{\theta_{\text{old}}}}
\!\left[1-r(\mathbf{x},\mathbf{c},\mathbf{P})\right] }.
\end{aligned}
\end{equation}
Policy improvement moves probability mass from $\pi^{-}$ toward $\pi^{+}$ while staying close to $\pi_{\theta_{\text{old}}}$, realizing via KL-regularized maximum likelihood.
Denoting $J(\theta,\mathbf{c},\mathbf{P}):=\mathbb{E}_{\mathbf{x}\sim \pi_{\theta}}\!\left[r(\mathbf{x},\mathbf{c},\mathbf{P})\right]$,
the updated policy $\pi_{\theta_{\text{new}}}$ satisfies the improvement criterion:
\begin{equation}
\Delta J \;:=\; J(\theta_{\text{new}},\mathbf{c},\mathbf{P})
- J(\theta_{\text{old}},\mathbf{c},\mathbf{P}) \;>\; 0 .
\end{equation}

\subsection{Verifiable Geometry Reward Design}
\label{sec:rwd}
For each generated video and its reference, a large 3D model is applied to estimate per-frame \emph{camera-to-world} extrinsics and per-pixel confidence maps, allowing us to derive the camera trajectory for both sides:
\begin{equation}
\begin{aligned}
&\mathbf{E}_n=[\mathbf{R}_n\mid \mathbf{t}_n],\quad
\tilde{\mathbf{E}}_n=[\tilde{\mathbf{R}}_n\mid \tilde{\mathbf{t}}_n],\\
&\tilde{\omega}_n=\mathrm{Agg}(\tilde{\Omega}_n)\in[0,1],\quad
\tilde{\bm{\Omega}}_n\in\mathbb{R}^{h\times w}.
\end{aligned}
\end{equation}
Here $\mathbf{E}$ and $\tilde{\mathbf{E}}$ denote extrinsics of the reference and generated video, $\tilde{\omega}$ denotes per-frame confidence of the generated video, and $\mathrm{Agg}(\cdot)$ is the operation of mean pooling.
To address the scale ambiguity, we first estimate an Umeyama transform $(s^\star,\mathbf{R}^\star,\mathbf{t}^\star)=\mathrm{Umeyama}\big(\{\tilde{\mathbf{o}}_n\}_{n=1}^{N},\{\hat{\mathbf{o}}_n\}_{n=1}^{N}\big)$ with two trajectories~\cite{lawrence2019purely}. Then, we apply that to each camera pose of the generated videos:
\begin{equation}
\tilde{\mathbf{E}}_n' 
= \big[\, \mathbf{R}^\star \tilde{\mathbf{R}}_n \;\big|\; s^\star \mathbf{R}^\star \tilde{\mathbf{t}}_n + \mathbf{t}^\star \,\big].
\end{equation}
This normalization helps alleviate the scale issue caused by the randomness of generated content, making the following reward calculation more convincing.

Afterward, we partition both trajectories into non-overlapping segments of length $L$. Within the $k$-th segment, we compute end-to-start relative transforms, emphasizing local smoothness:
\begin{equation}
\tilde{\mathbf{T}}_k=(\tilde{\mathbf{E}}_{n_k}')^{-1}\tilde{\mathbf{E}}_{n_k+L}',\quad
\hat{\mathbf{T}}_k=(\hat{\mathbf{E}}_{n_k})^{-1}\hat{\mathbf{E}}_{n_k+L}.
\end{equation}
We form the error transform $\mathbf{T}^{\text{err}}_k=\hat{\mathbf{T}}_k^{-1}\tilde{\mathbf{T}}_k$ and extract translation component $\mathbf{R}^{\text{err}}_k=\mathbf{R}(\mathbf{T}^{\text{err}}_k)$ and rotation component $\mathbf{t}^{\text{err}}_k=\mathbf{t}(\mathbf{T}^{\text{err}}_k)$. Subsequently, the per-segment errors are calculated as:
\begin{equation}
\begin{aligned}
e_t(k)&=\mathrm{clip}\left(\left\|\mathbf{t}^{\text{err}}_k\right\|_2,\,-1,\,1\right),\\
e_R(k)&=\arccos\Big(\mathrm{clip}\big(\tfrac{\mathrm{tr}(\mathbf{R}^{\text{err}}_k)-1}{2},-1,1\big)\Big).
\end{aligned}
\end{equation}
Errors are converted to alignment scores $\mathbf{s}$ by simple reversal,
\begin{equation}
\mathbf{s}_k \;=\; -\big(\lambda_t\, e_t(k)+\lambda_R\, e_R(k)\big),
\end{equation}
where $\lambda_t$ and $\lambda_R$ are hyper-parameters to balance the importance of translation and rotation.
In doing so, better-aligned segments (smaller errors) yield larger $\mathbf{s}_k$ (less negative). Given the per-frame confidence, we further mask out unreliable segments via $\mathbf{m}_k=\mathbf{1}\!\left[\bm{\omega}_k\ge \tau\right]$ and obtain the final reward $\tilde{\mathbf{s}}_k$. This construction yields dense feedback that is insensitive to the ambiguous scale, locally descriptive, and compatible with online RL.

\subsection{Training Paradigm}
\label{sec:train}
By resorting to our verifiable geometry reward function, we conduct online RL training on the video diffusion model $\pi_{\theta}$ with group-relative policy optimization (GRPO)~\cite{liu2025flow,shao2024deepseekmath,xue2025dancegrpo}. As shown in Figure~\ref{fig:method}, for each pair of conditions $(\mathbf{c},\mathbf{P})$, we sample a group of $G$ rollouts under a stochastic reverse process so that each per-step policy is Gaussian,
$\pi_{\theta}(\mathbf{x}_{t-1}\mid\mathbf{x}_t,\mathbf{c},\mathbf{P})=\mathcal{N}\big(\boldsymbol{\mu}_{\theta}(\mathbf{x}_t,t,\mathbf{c},\mathbf{P}),\,\sigma_t^2\mathbf{I}\big)$, thereby injecting exploration noise while keeping the original marginals. The reward function assigns a list of reward values to the $g$-th rollout $\tilde{\mathbf{s}}_{1:K}^{g}=\{\tilde{\mathrm{s}}_k^{g}\}_{k=1}^{K}$. With a group of reward values, we adopt a ``z-score'' normalization to obtain group-relative advantages below:
\begin{equation}
\mathbf{A}_k^{g}=\big({\mathrm{s}}_{k}^{g}-\mu\big)\big/\big(\delta+\gamma\big),
\end{equation}
where $\mu$ and $\delta$ denote the mean and standard deviation of all reward values in a group. Here $\gamma$ is used to improve numerical stability. Afterward, we optimize the video diffusion model $\pi_{\theta}$ by maximizing the following objective:
\begin{equation}
\max_{\theta}\frac{1}{G\,T\,K}\sum_{g=1}^{G}\sum_{t=1}^{T}\sum_{k=1}^{K}
\Big(
f\big(\mathbf{r}^g_t,\mathbf{A}^g_k,\Delta\big)-\beta\,\mathrm{KL}(\pi_{\theta}\,\|\,\pi_{\mathrm{ref}})
\Big),
\end{equation}
where the ratio $\mathbf{r}^g_t$ is computed by $\frac{p_{\theta}(\mathbf{x}^{g}_{t-1}\mid \mathbf{x}^{g}_{t},\mathbf{c},\mathbf{P})}{p_{\theta_{\text{old}}}(\mathbf{x}^{g}_{t-1}\mid\mathbf{x}^{g}_{t},\mathbf{c},\mathbf{P})}$, and $f(\cdot)$ denotes $\min\!\big(\mathbf{r}^{g}_t\,\mathbf{A}^{g}_k,\;\mathrm{clip}(\mathbf{r}^{g}_t,1-\Delta,1+\Delta)\,\mathbf{A}^{g}_k\big)$. We initialize \(\pi_{\mathrm{ref}}\) by the model after supervised fine-tuning, which can mitigate reward hacking. In practice, we use a few-step scheduler to collect rollouts efficiently.

\section{Experiments}
\label{sec:exp}

\subsection{Implementation Details}
\label{sec:imple}
We adopt the latent diffusion architecture~\cite{rombach2022high}, including a spatial-temporal variation autoencoder (ST-VAE) and a diffusion transformer. The input video is downsampled by 4 for temporal and 8 for spatial via ST-VAE. The diffusion transformer is initialized by an internal pre-trained model, with about 2B parameters. We interpolate input camera poses to the same number of frames in the video data, because only keyframes are annotated with camera information. These interpolated camera poses are further embedded by a light-weight camera network, as the input camera condition.

\myPara{Supervised fine-tuning.} 
We conduct supervised fine-tuning on a curated dataset consisting of 315k text-video-camera samples, with 314k used for training and 1k reserved for evaluation. Each video is preprocessed by resizing the longer side to 512 pixels while preserving the aspect ratio. Video durations range from 2 to 10 seconds.
Our infrastructure supports variable-length sequences. We use a packing dataloader to maintain load balance during training. The training is performed with a batch size of approximately 128, and all model parameters are fine-tuned for 10k iterations. We apply a text condition dropout ratio of 0.3 during training, while the camera condition is always retained to learn camera movement.
Training is conducted in a multi-task setting, incorporating both text-to-video and image-to-video generation. Task sampling is imbalanced, with 30\% text-to-video and 70\% image-to-video.
We employ the AdamW optimizer with a learning rate of $5 \times 10^{-5}$, weight decay of 0.01, and $\epsilon$ set to $1 \times 10^{-15}$. To effectively learning camera movement, the timestep scheduler is shifted by 5. Note that, for efficiency, we pre-extract both the VAE latent representations and the caption embeddings before training begins.

\myPara{GRPO fine-tuning.} 
We perform GRPO-based fine-tuning on the image-to-video task, using identical hyperparameters for both. A subset of approximately 3.2k samples is selected from the training set for this stage.
We apply LoRA~\cite{hu2022lora} to all linear layers within each transformer block, with a rank of $r=64$ and $\alpha=128$. Fine-tuning is conducted for 200 iterations.
Each training group consists of 16 samples. From each group, we select the top 4 and bottom 4 samples based on reward values for GRPO optimization.
We use 14 timesteps for the denoising process. The first three steps adopt a stochastic differential equation (SDE) sampler with a noise level of 0.7, while the remaining steps utilize a first-order ordinary differential equation~(ODE)-based sampler. The timestep scheduler is shifted by 6.
We set the classifier-free guidance (CFG) scale to 3.5 and apply a KL divergence loss with a weight of $1 \times 10^{-4}$ to avoid reward hacking. All experiments are conducted on 32$\times$NVIDIA H200 GPUs.

\begin{table*}[!t]
    \centering
    \caption{
        \textbf{Quantitative comparison with state-of-the-arts.}
        We evaluate each model using metrics of camera control accuracy (\ie, ``\textit{Trans. Err.}'' and ``\textit{Rot. Err.}''), geometric consistency (\ie, ``\textit{Geo. Con.}'') and visual quality (\ie, ``\textit{VQ}'') on the RealEstate10K test set.
        For CameraCtrl~\cite{cameractrl}, we evaluate the ADv3-based model for the text-to-video~(T2V) setting and the SVD-based model for the image-to-video~(I2V) setting.
        The released weight of AC3D~\cite{ac3d} only supports text-to-video generation. ``\method$^{*}$'' is only optimized by supervised fine-tuning. The best result in each case is marked in bold.
    }
    \vspace{-5pt}
    \label{tab:tab1}
    \setlength{\tabcolsep}{2.2mm}{
    \begin{tabular}{l|cccc|cccc}
    \toprule
     \multirow{2}[2]{*}{\textbf{Method}} & \multicolumn{4}{c|}{\textbf{Text-to-Video}} & \multicolumn{4}{c}{\textbf{Image-to-Video}} \\
     \cmidrule(lr){2-5}\cmidrule(lr){6-9}
     ~ & \textit{Trans. Err.}$\downarrow$ & \textit{Rot. Err.}$\downarrow$ & \textit{Geo. Con.}$\uparrow$ & \textit{VQ}$\uparrow$ & \textit{Trans. Err.}$\downarrow$ & \textit{Rot. Err.}$\downarrow$ & \textit{Geo. Con.}$\uparrow$ & \textit{VQ}$\uparrow$ \\
    \midrule
    CameraCtrl~\cite{cameractrl} & 0.0887 & 1.4586 & 0.7567 & 2.93 & 0.1696 & 3.2809 & 0.6087 & 2.63 \\
    AC3D-2B~\cite{ac3d} & 0.0476 & 1.0451 & 0.8156 & 3.82 & - & - & - & - \\
    AC3D-5B~\cite{ac3d} & 0.0428 & 0.9120 & 0.8820 & 4.69 & - & - & - & - \\
    \midrule
    \method$^{*}$ & 0.0395 & 0.6506 & 0.9081 & 5.30 & 0.0337 & 0.5613 & 0.9174 & \textbf{5.02} \\
    \method & \textbf{0.0293} & \textbf{0.5140} & \textbf{0.9173} & \textbf{5.91} & \textbf{0.0286} & \textbf{0.4685} & \textbf{0.9226} & 4.87 \\
    \bottomrule
    \end{tabular}}
    \vspace{-5pt}
\end{table*}

\subsection{Evaluation}
\label{sec:eval}
\myPara{Datasets.} The evaluation is conducted on a test set consisting of 1k video clips. This set includes 342 clips from the RealEstate10K dataset~\cite{realestate10k} and 658 clips from our curated dataset. These clips are sampled to cover a diverse range of camera trajectories, ensuring a comprehensive assessment of generation quality and robustness. All video durations are constrained between 4 and 8 seconds.
For the ablation study, we randomly select a subset of 100 clips from the 342 RealEstate10K samples to accelerate the evaluation process while preserving effectiveness.

\myPara{Metrics.} 
Our evaluation framework assesses model performance across three key aspects: visual quality, camera control accuracy, and geometric consistency. Specifically, 
\begin{itemize}
    \item Camera control accuracy: We assess camera trajectory alignment by measuring both translation and rotation errors between the generated and reference videos. For each input video, we use $\pi^3$~\cite{wang2025pi} to estimate camera trajectories. To account for scale ambiguity, we align the estimated trajectory of the generated video to that of the reference video using a similarity transformation. The translation error is computed as the average RMSE between corresponding camera positions, while the rotation error is defined as the mean angular difference between corresponding orientations.
    \item Geometric consistency: In addition to trajectory estimation, $\pi^3$~\cite{wang2025pi} outputs a per-pixel confidence map. We threshold this confidence map to obtain a binary mask, and define the geometric consistency score as the ratio of valid pixels to the total number of pixels. Here, the threshold is set to 0.1.
    \item Visual quality: To evaluate visual fidelity, we uniformly sample 8 frames from each generated video clip and compute the HPSv3 score using the prompt ``\texttt{A high-quality, clear video frame.}'' This metric quantitatively reflects perceptual quality as judged by a powerful vision-language model.
\end{itemize}

\begin{figure*}
    \centering
    \includegraphics[width=1\linewidth]{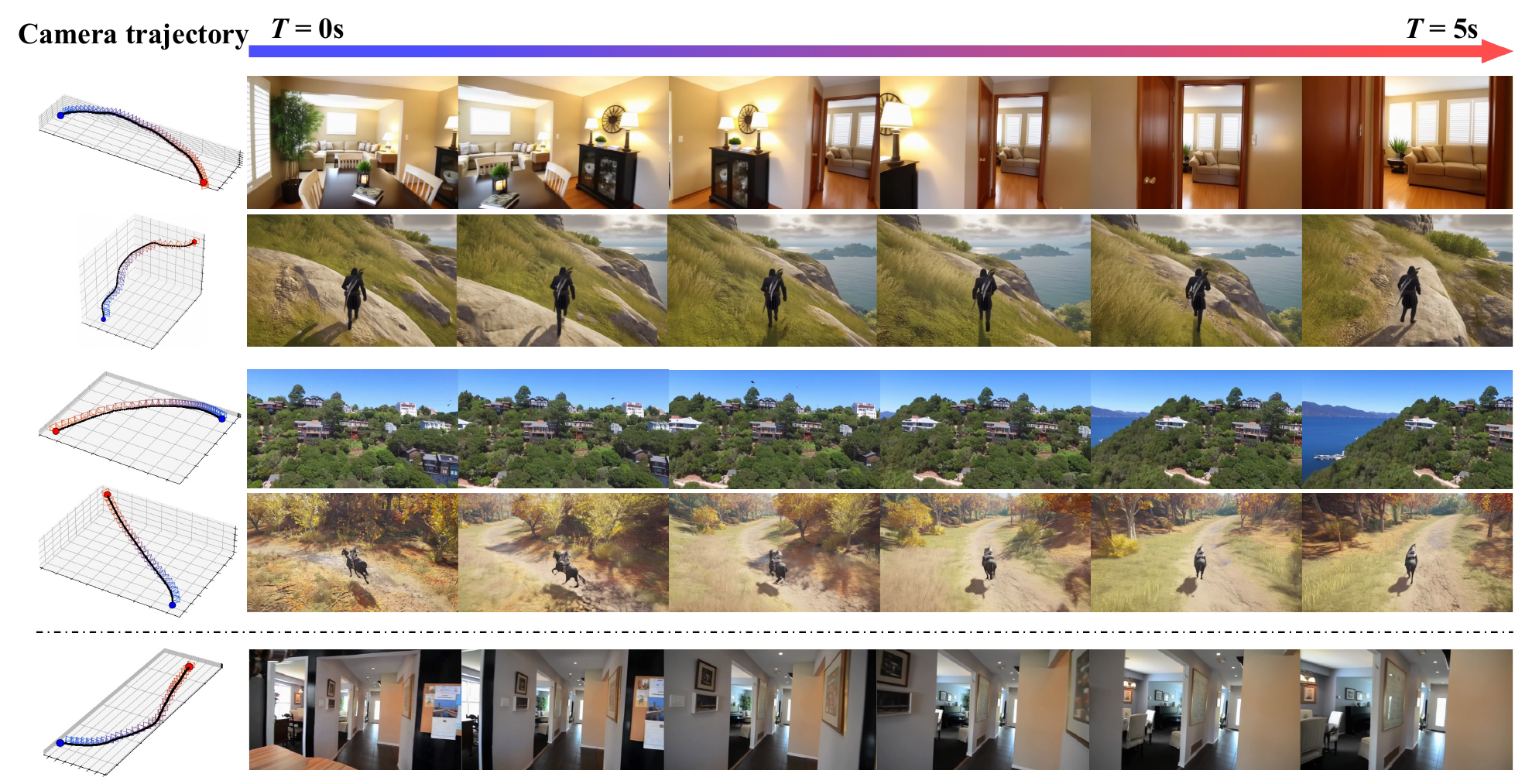}
    \caption{
    \textbf{Illustrations of qualitative results.} For each example, we condition on a reference camera trajectory (start in blue, end in red) and show uniformly sampled frames from $T=0$ to $T=5$s. The first and second rows are text-to-video~(T2V) cases. The third and fourth rows are image-to-video~(I2V) cases. The last row is a failure case. In the first half of the clip, the camera does not execute right-forward translation and instead moves straight forward, resulting in early trajectory misalignment.
    }
    \label{fig:vis}
\vspace{-0.2cm}
\end{figure*}
\begin{table*}[!t]
    \centering
    \caption{
        \textbf{Quantitative results of supervised fine-tuning.}
        We evaluate each model using metrics of camera control accuracy (\ie, ``\textit{Trans. Err.}'' and ``\textit{Rot. Err.}'') and visual quality (\ie, ``\textit{VQ}'') on the RealEstate10K test set.
        ``\method$^{*}$'' adopts a training recipe consisting of 3:7 T2V-I2V tasks by default.
        Results are reported for both text-to-video and image-to-video settings. The best result in each case is marked in bold.
    }
    \vspace{-5pt}
    \label{tab:tab2}
    \setlength{\tabcolsep}{4.5mm}{
    \begin{tabular}{l|ccc|ccc}
    \toprule
     \multirow{2}[2]{*}{\textbf{Method}} & \multicolumn{3}{c|}{\textbf{Text-to-Video}} & \multicolumn{3}{c}{\textbf{Image-to-Video}} \\
     \cmidrule(lr){2-4}\cmidrule(lr){5-7}
     ~ & \textit{Trans. Err.}$\downarrow$ & \textit{Rot. Err.}$\downarrow$ & \textit{VQ}$\uparrow$ & \textit{Trans. Err.}$\downarrow$ & \textit{Rot. Err.}$\downarrow$ & \textit{VQ}$\uparrow$ \\
    \midrule
    \rowcolor{ourblue!50}
    \method$^{*}$               & \textbf{0.0395} & \textbf{0.6506} & 5.30 & \textbf{0.0337} & \textbf{0.5613} & 5.02 \\
    \textit{w/} 7:3 T2V-I2V         & 0.0411 & 0.7112 & 5.42 & 0.0363 & 0.6378 & 5.05 \\
    \textit{w/o} dynamic content    & 0.0464 & 0.7306 & \textbf{5.45} & 0.0394 & 0.6555 & \textbf{5.35} \\
    \textit{w/o} shifted timesteps  & 0.0420 & 0.7023 & 5.12 & 0.0342 & 0.6055 & 4.96 \\
    \textit{w/o} pose normalization & 0.0416 & 0.7458 & 5.18 & 0.0356 & 0.7135 & 4.99 \\
    \bottomrule
    \end{tabular}}
\vspace{-1.5em}
\end{table*}

\begin{table}[!t]
    \centering
    \caption{
        \textbf{Effects on different combinations of error functions.}
        We evaluate each setting with metrics of camera control accuracy (\ie, ``\textit{Trans. Err.}'' and ``\textit{Rot. Err.}'') on the RealEstate10K test set (100 clips).
        $e_{t}(\cdot)$ and $e_{R}(\cdot)$ indicate the error function of translation and rotation, respectively.
        We report results on the image-to-video setting.
        $^{\dagger}$ denotes the clip-level reward. The best result in each case is marked in bold.
    }
    \label{tab:tab3}
    \setlength{\tabcolsep}{4.4mm}{
    \begin{tabular}{cc|cc}
    \toprule
    $e_{t}(\cdot)$ & $e_{R}(\cdot)$ & \textit{Trans. Err.}$\downarrow$ & \textit{Rot. Err.}$\downarrow$ \\
    \midrule
    \rowcolor{ourblue!50}
    \multicolumn{2}{c|}{SFT baseline} & 0.0280 & 0.4444 \\
    \textit{rel.} & \textit{abs.} & 0.0292 & 0.4537 \\
    \textit{abs.} & \textit{rel.} & 0.0293 & 0.4281 \\
    \textit{rel.} & \textit{rel.} & \textbf{0.0238} & \textbf{0.3286} \\
    \textit{rel.}$^{\dagger}$ & \textit{rel.}$^{\dagger}$ & 0.0277 & 0.3739 \\
    \bottomrule
    \end{tabular}}
\end{table}

\begin{table}[!t]
    \centering
    \caption{
        \textbf{Effects on different numbers of timesteps.}
        Given the fixed number of sampling steps $T_{\text{total}}$, we evaluate effects on different numbers of SDE steps $T_{\text{sde}}$ used in optimization.
        ``\textit{Time.}'' denotes the consumed time per optimization step.
        We report results on the image-to-video setting. The best result in each case is marked in bold.
    }
    \label{tab:tab4}
    \setlength{\tabcolsep}{3.2mm}{
    \begin{tabular}{c|ccc}
    \toprule
    $T_{\text{sde}}$ ($T_{\text{total}}$) & \textit{Trans. Err.}$\downarrow$ & \textit{Rot. Err.}$\downarrow$ &\textit{Time.}$\downarrow$ \\
    \midrule
    1 (14) & 0.0272 & 0.4829 & \textbf{274}s \\
    3 (14) & \textbf{0.0238} & \textbf{0.3286} & 343s \\
    5 (14) & 0.0278 & 0.4440 & 412s \\
    \bottomrule
    \end{tabular}}
\end{table}

\begin{table}[!t]
    \centering
    \caption{
        \textbf{Effects on best-of-$N$ sampling.}
        We evaluate effects on different numbers of positive and negative samples.
        ``$\alpha$-$\beta$'' indicates $\alpha$ positive and $\beta$ negative samples.
        ``\textit{Time.}'' denotes the consumed time per optimization step.
        We report results on the image-to-video setting. The best result in each case is marked in bold.
    }
    \label{tab:tab5}
    \setlength{\tabcolsep}{3.3mm}{
    \begin{tabular}{c|ccc}
    \toprule
    Best-of-$N$ & \textit{Trans. Err.}$\downarrow$ & \textit{Rot. Err.}$\downarrow$ &\textit{Time.}$\downarrow$ \\
    \midrule
    4-4 & \textbf{0.0238} & \textbf{0.3286} & 343s \\
    2-2 & 0.0240 & 0.3760 & 294s \\
    1-1 & 0.0249 & 0.4330 & \textbf{269}s \\
    \midrule
    1-4 & 0.0303 & 0.4948 & 305s \\
    4-1 & 0.0262 & 0.3487 & 305s \\
    \bottomrule
    \end{tabular}}
\end{table}

\begin{table}[!t]
    \centering
    \caption{
        \textbf{Effects on different ranks of LoRA.}
        We evaluate each setting with metrics of camera control accuracy (\ie, ``\textit{Trans. Err.}'' and ``\textit{Rot. Err.}'') on the RealEstate10K test set (100 clips).
        We report results on the image-to-video setting. The best result in each case is marked in bold.
    }
    \label{tab:tab6}
    \setlength{\tabcolsep}{6.8mm}{
    \begin{tabular}{c|cc}
    \toprule
    Rank $r$ & \textit{Trans. Err.}$\downarrow$ & \textit{Rot. Err.}$\downarrow$ \\
    \midrule
    32 & \textbf{0.0233} & 0.3550 \\
    64 & 0.0238 & \textbf{0.3286} \\
    128 & 0.0249 & 0.3778 \\
    \bottomrule
    \end{tabular}}
\end{table}

\begin{table}[!t]
    \centering
    \caption{
        \textbf{Effects on different timestep shift.}
        We evaluate each setting with metrics of camera control accuracy (\ie, ``\textit{Trans. Err.}'' and ``\textit{Rot. Err.}'') on the RealEstate10K test set (100 clips).
        We report results on the image-to-video setting. The best result in each case is marked in bold.
    }
    \label{tab:tab7}
    \setlength{\tabcolsep}{5mm}{
    \begin{tabular}{c|cc}
    \toprule
    Timestep shift & \textit{Trans. Err.}$\downarrow$ & \textit{Rot. Err.}$\downarrow$ \\
    \midrule
    4 & 0.0264 & 0.3878 \\
    6 & \textbf{0.0238} & \textbf{0.3286} \\
    8 & 0.0297 & 0.4165 \\
    \bottomrule
    \end{tabular}}
\vspace{-0.3cm}
\end{table}

\subsection{Main Results}
\label{sec:main_rst}

\myPara{Quantitative comparison.}
As demonstrated in Table~\ref{tab:tab1}, we compare our method with other state-of-the-art methods on the RealEstate10K test set, using camera-control accuracy, geometric consistency, and visual quality (VQ) in both text-to-video (T2V) and image-to-video (I2V) settings. Benefiting from the advanced training recipe and our curated dataset, \method$^*$ clearly surpasses existing methods~\cite{ac3d,cameractrl} by large margins. Despite our strong SFT baselines, our online RL post-training (see \method) further improves camera control and geometry. In T2V generation, the translation error drops \textbf{25.8\%} from 0.0395 to 0.0293, while the rotation error drops \textbf{21.0\%} from 0.6506 to 0.5140. Geometric consistency rises from 0.9081 to 0.9173 after post-training.
Similar trends also show in I2V generation. Concretely, translation and rotation errors decrease by \textbf{15.1\%} and \textbf{16.5\%}, respectively.
In the I2V setting, VQ slightly decreases after post-training (from 5.02 to 4.87), consistent with our reward not directly optimizing perceptual quality. Incorporating a multi-objective reward is an effective way to alleviate this issue.

\myPara{Qualitative analysis.}
We visualize results for both T2V and I2V settings in Figure~\ref{fig:vis}. Across a wide range of diverse, large-amplitude trajectories, our model closely follows the reference camera trajectories and maintains smooth local viewpoint changes at turning points. Meanwhile, subjects exhibit substantial dynamics (\eg, running and riding a horse), and the rendered scenes remain temporally coherent, indicating that the model can disentangle camera movement from content motion while preserving visual fidelity. For the shown failure, straightforward translation is still the dominant pattern in the dataset, even diversifies the trajectory distribution. The model therefore drifts toward straightforward motion rather than exactly executing the reference trajectory, resulting in early misalignment. Mitigating the dataset bias is an important direction for the research community.

\subsection{Ablation Study}
\label{sec:ablation}
We analyze 6 design choices that influence GRPO fine-tuning. Unless otherwise noted, all of the results are reported on the RealEstate10K test set~\cite{realestate10k}, containing 100 clips, in the I2V setting. We mainly evaluate camera control accuracy using translation error (\textit{Trans. Err.}) and rotation error (\textit{Rot. Err.}). For training efficiency, we report the time spent on the optimization step.

\myPara{SFT training recipe.} As illustrated in Table~\ref{tab:tab2}, we first compare different sample ratios for T2V and I2V, demonstrating that animation of the first frame encourages the model to learn camera movement. Besides, training without dynamic videos clearly degrades camera control accuracy. Afterwards, the shifted timestep scheduler and pose normalization each contribute to advancing camera-control video generation. We therefore adopt this recipe for all RL experiments, excluding the sample ratio for T2V and I2V.

\myPara{Error function.} It is crucial to design an appropriate error function for the reward model. As reported in Table~\ref{tab:tab3}, using the relative error for both translation and rotation yields the lowest overall errors, outperforming the other mixed configurations. We further replace the dense, segment-level reward with a clip-level reward and apply GRPO fine-tuning. This variant results in noticeably worse camera-control accuracy, demonstrating the importance of dense feedback.

\myPara{Number of SDE timesteps.}
With total sampling steps fixed, we vary the number of stochastic steps $T_{\text{sde}}$ used during online rollouts, shown in Table~\ref{tab:tab4}. $T_{\text{sde}}=3$ shows the most accurate control with 0.0238 translation error and 0.3286 rotation error at a reasonable cost. Too few steps are under-explored, while too many steps slow training and slightly decrease accuracy.

\myPara{Best-of-$N$ sampling.}
We evaluate the groups with $\alpha$ positive and $\beta$ negative samples. A balanced larger group ($4\!-\!4$) achieves the best performance of 0.0238 translation error and 0.3286 rotation error. Smaller or asymmetric groups (\eg, $1\!-\!1$ and $1\!-\!4$) increase both translation and rotation errors, which indicate that both adequate positive and negative samples are important.

\myPara{Rank in LoRA.}  
In Table~\ref{tab:tab6}, we evaluate the effects of capacity on different ranks in LoRA. As shown, $r=64$ offers the best overall trade-off between camera control accuracy and trainable parameters. $r=32$ slightly decreases translation error but harms rotation accuracy, while $r=128$ degrades control accuracy and increases training costs.

\myPara{Timestep shift.}  
In the GRPO fine-tuning stage, we offset the diffusion timesteps by different constants when sampling and policy optimization, shown in Table~\ref{tab:tab7}. Camera control accuracy is highly sensitive to the model’s behavior near the start of denoising. If the shift is too small, these noise-proximal steps are under-trained, and errors accumulate along the denoising process. Conversely, excessive shift overpowers extremely noisy states, producing blurrier rollouts. This causes challenges to camera pose estimation, resulting in inferior camera control accuracy.

\section{Conclusion}
\label{sec:conclusion}
In this paper, we presented \method, an online RL framework that post-trains video diffusion models for precise camera control. Through a verifiable geometry reward that converts 3D alignment into dense segment-level feedback, our method effectively mitigates reward sparsity and leads to efficient model optimization. Coupled with a large-scale dataset covering diverse and high-amplitude camera trajectories, \method achieves significant improvements in camera-control accuracy, geometric consistency, and visual quality over SFT competitors. These results highlight the promise of RL as a powerful post-training paradigm for refining generative video models beyond static supervision. In future work, we will integrate our method into the world model for precise action control and explore that as a scalable data engine for embodied AI.
{
    \small
    \bibliographystyle{ieeenat_fullname}
    \bibliography{main}
}

\end{document}